# Self-Supervised-RCNN for Medical Image Segmentation with Limited Data Annotation

Banafshe Felfeliyan[1], Abhilash Hareendranathan[2], Gregor Kuntze[1], David Cornell[2], Nils D. Forkert[1], Jacob L. Jaremko[23], and Janet L. Ronsky[1]

[1] University of Calgary, Calgary, Alberta, Canada
[2] University of Alberta, Edmonton, Alberta, Canada
[3] Alberta Machine Intelligence Institute (AMII), Edmonton, Alberta, Canada
`email@domain.com`

**Abstract.** Many successful methods developed for medical image analysis that are based on machine learning use supervised learning approaches, which often require large datasets annotated by experts to achieve high accuracy. However, medical data annotation is time-consuming and expensive, especially for segmentation tasks. To solve the problem of learning with limited labeled medical image data, an alternative deep learning training strategy based on self-supervised pretraining on unlabeled MRI scans is proposed in this work. Our pretraining approach first, randomly applies different distortions to random areas of unlabeled images and then predicts the type of distortions and loss of information. To this aim, an improved version of Mask-RCNN architecture has been adapted to localize the distortion location and recover the original image pixels. The effectiveness of the proposed method for segmentation tasks in different pre-training and fine-tuning scenarios is evaluated based on the Osteoarthritis Initiative dataset. Using this self-supervised pretraining method improved the Dice score by 20% compared to training from scratch. The proposed self-supervised learning is simple, effective, and suitable for different ranges of medical image analysis tasks including anomaly detection, segmentation, and classification.

## 1 Introduction

Accurate quantification of image-based biomarkers from medical images related to pathologies and inflammation are needed for an improved clinical management and for identifying novel targets for therapy. With the development of deep learning techniques, an increased success in the automation of radiological assessments can be observed [7]. Many of these techniques rely on supervised learning approaches, which often require large, accurate annotated datasets is key to their success. However, acquiring "ground truth" labels for large datasets to train deep learning networks is expensive and time-consuming while learning from limited labeled data remains a fundamental problem. This requirement becomes even more complicated and challenging when the aim is to identify medical image features and pathologies that are more complex and require a higher level of expertise to generate ground truth data. For example, for effusion segmentation task in knee MRI scans, pockets of joint fluid (effusion) can exist in unexpected locations, and more expertise is needed to distinguish effusion from other fluids, and minimize risk of inattentional error [2]. Furthermore, the annotation cost increases exponentially with semantic segmentation, which requires pixel-by-pixel labeling. To solve the problem of training with limited or no data annotation different strategies have been reported including domain adaptation [8], data generation and augmentation [3], knowledge distillation [28] and few-shot learning [22].

Another approach to train a network with limited labeled data is using pre-trained networks and finetuning its parameters for the target task using limited labeled data. Two common pretraining approaches for this purpose include: (1) supervised pretraining on a large labeled dataset (e.g., ImageNet [4] and MS-COCO [18]), and (2) self-supervised learning (SSL) pretraining on unlabeled data.

Due to the large domain gap between natural images and medical images, the networks pretrained using natural image improve performance in the medical imaging domain [24]. Therefore recently, to learn more relevant representations, more of the medical image analysis research has moved towards SSL which is useful to pretrain networks with unlabeled domain-specific images. More precisely, SSL first learns features from unlabeled datasets usually with pretext tasks (pre-designed tasks for networks to solve), and then transfers the knowledge to target tasks (Fig. 1. (a)). One of the main challenges with SSL algorithms is defining pretext tasks in a manner that relates to the final applications so that relevant features are identified in this stage. Most of the current SSL methods learn image-level pretext tasks like predicting augmentations applied to an image or learning to discriminate between images [31]. However, as the segmentation task requires pixel-level prediction, a self-supervised dense representation learning will be more suitable for segmentation tasks.

To achieve this goal, we propose a self-supervised pretraining using the Mask R-CNN (SS-MRCNN) approach for medical image segmentation with limited labels. In this method the network learns to perform three pretext tasks 1) localize distortions, 2) classify distortions type, and 3) recover the distorted area. During the pretraining different distortions are applied to random areas of the input images and the network is trained to recover the lost information, ensuring the network is receiving different inputs at each epoch even with few data points. Using this principle, SS-MRCNN can be trained with both small and large datasets. We demonstrate training on proposed pretext tasks provides a powerful objective for SSL pretraining to learn different levels of image features. We perform a comparison between self-supervised and supervised pretraining on a musculoskeletal (MSK) image segmentation task. Key contributions of this research include: 1) proposing a new simple multi-task SSL approach using Mask RCNN, 2) investigating the benefit of SSL pretraining on medical image segmentation task with limited labeled data, 3) demonstrating that the proposed SSL pre-training is robust and generalizable for different training datasets sizes.

## 2 Related Work

### 2.1 Transfer Learning for Medical Image Analysis

Many previous studies have taken advantage of transfer learning strategies for medical image analysis. The majority of transfer learning methods use pretrained standard ImageNet architectures (e.g. ResNet) and fine-tune network weights based on the target task [19][21][14][29]. However, as Raghu et al. [24] demonstrated while this strategy can make convergence faster it may not provide significant improvement in medical image classification, mainly attributable to the domain gap between natural images and medical images. To alleviate this problem some previously described methods normalized medical input images based on ImageNet statistics [20]. However, this approach is usually not practical for radiological images since they are grayscale images and have completely different data distribution.
Using domain data provides an alternate solution for pre-trained networks in the medical domain [30][17][9]. Roy et. al. used auxiliary labels created from segmentation software for pretraining [9]. Karimi et. al. showed [16], as the size of the target dataset is reduced, transfer learning with medical domain data can have more positive effect. However, obtaining large labeled datasets in the same domain often is not feasible. In addition, the positive effects of transfer learning depend on similarity between the pretext and target tasks. He et al. [12] showed pretraining with classification tasks (e.g. ImageNet pre-training) has no benefit for target tasks that are sensitive to spatially well-localized data like segmentation.

### 2.2 Self-supervised Learning

In SSL, image representations are acquired directly from the image pixels, without relying on semantic annotations, typically through learning pretext tasks such as applying a transformation and determining the transformation type from the transformed image. A systematic analysis in medical imaging domain showed that SSL models outperform the models that make use of ImageNet supervised pretraining [13].

Contrastive-Learning (CL) is one of the popular SSL methods, that provides a decent initialization to fine-tune a target task, particularly when limited annotations are available [28]. The pretext task in CL is instance discrimination, with the objective to maximize the similarity between similar pairs (positive) and minimize the similarity between dissimilar (negative) pairs [25][28]. Within the medical image analysis domain, CL was successfully used along with federated learning for medical image segmentation [27]. Azizi et al. [1] proposed a multi-instance CL (MICL) framework pre-training for medical image classification. Nevertheless, CL algorithms have some issues, including the challenge of choosing dissimilar pairs, which is critical to the quality of learned representations [28], the need for large amounts of memory to retain negative pairs, and the fact that they are designed for large-scale datasets with diversity [27].

An alternative approach to CL is to mask some information of the input and define training the objective as recovering the original data. Some SSL methods use image inpainting and recovering corruption as the pretext tasks [23][5][15][26]. Feature learning with context encoders and inpainting as pretext was one of the early SSL methods [23]. Wei et al. [26] proposed an SSL pretraining with the objective of predicting the Histograms of Oriented Gradients (HOG) of corrupted areas of the image using transformers. These methods are conceptually and practically simpler than CL-based methods and more preferable for learning dense image representation.

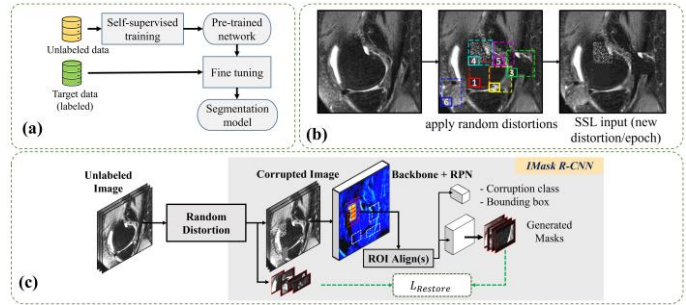

**Fig. 1.** (a) SSL paradigm, (b) Dataflow of distortion block and different distortions: 1-blank, 2-blurred, 3-mislocate, 4-salt&pepper noise, 5-rotate, 6-speckle noise, (c) proposed SSL pipeline

## 3  Method

The proposed SS-MRCNN framework uses an improved version of the Mask RCNN (IMaskRCNN [6]) architecture (Fig. 1). Similar to the Mask RCNN [11], the IMaskRCNN is constructed from a backbone (ResNet) for feature extraction, a region proposal network (RPN) for extracting the ROI bounding box, and two heads: the mask head for mask segmentation, and the classification head for extracted bounding box classification. Compared to the Mask R-CNN, pixel-wise prediction is improved in IMaskRCNN by adding a skip connection and an extra encoder layer to the mask head.

### 3.1  Self-Supervised Pretraining

The IMaskRCNN [6] is adapted to perform pretext tasks, it receives distorted image as input (Fig. 1(b)) for the SSL training. The main objective of the SSL pretraining is to learn dense and semantic representation of the input image by three pretext tasks: 1) localize the distorted area, 2) identify distortions type, and 3) recover distorted areas.

**Distortion block:** The basic idea of the method proposed in this work is that in each iteration a random number of random distortions (3 to 7) is applied to randomly selected areas of the unlabeled image. The distortions are selected from the specified distortion pool ("blank", "blurred", "mislocate", "salt & pepper noise", "rotate", "speckle noise"). Identifying and recovering each distortion is equivalent to learning an image processing task and helps the network to learn different features regarding the input. For example, "blank" and "blurred" distortions are equivalent to image inpainting and super resolution tasks, respectively, which will be effective in learning the image structure and dense representation. With "mislocate" and "rotate", the network learns geometric information. By restoring information that has been lost by adding "salt & pepper noise" and "speckle noise", the network learns about the data distribution and structure.

**Loss:** The IMaskRCNN has a multitask learning loss (L) for each sampled ROI, which is the accumulation of the classification loss ($L_{cls}$), the bounding-box loss ($L_{bbox}$), and the mask-loss ($L_{mask}$) ($L = L_{cls} + L_{bbox} + L_{mask}$). For the SSL training, the $L_{mask}$ is replaced with the $L_{Restored}$, which is constructed from regression and similarity losses instead of Binary-cross-entropy. To measure the distance between the distribution of the real mask and the generated mask we defined $L_{Restored}$ as summation of RMSE (root means squared error), MAE (Mean Squared Error) and the Cosine Similarity (CS) loss, $L_{Restored} = L_{rmse} + L_{mae} + L_{CS}$.

## 4 Experiments and Results

In order to evaluate the added benefit of the proposed approach, we compare fine-tuning performance of using the proposed SSL pretraining (SS-MRCNN) to the network pretrained using MS-COCO dataset [18] (200K labeled images and 1.5 million instances) for the selected medical image segmentation tasks.

### 4.1 Implementation:

The proposed method was implemented in TensorFlow2 with Keras backbone and trained on a NVIDIA V100 GPU. For all experiments, input images were cropped/padded to 320 × 320 pixels. For the SSL pretraining models were trained for 200K iterations with batch size 128 ( =32 for small dataset), using the Adam optimizer and 0.001 learning rate. During the fine-tuning the model was trained for 100 epochs with a batch size of 16 using Adam optimizer and 0.001 learning rate. The weights from the pretrained network were used for the IMaskRCNN backbone for the segmentation task (target task), and only the IMaskRCNN heads were trained. Width and length of distortions were prescribed to be between 50 to 80 pixels.

### 4.2 Application and datasets:

For the experiments, we used seven different MRI sequences from the publicly available multicenter Osteoarthritis Initiative (OAI, https://nda.nih.gov/oai/) dataset. OAI contains thousands of MRI scans from 4796 subjects aged 45-79 years with 10 years annual knee assessments. Furthermore, to test generalizability of SS-MRCNN to unseen sequence from unseen body regions, we used 10 scans from a Clinical Hip Dataset that contains Short-TI Inversion Recovery (STIR) MRI scan (UofA HREB Pro00039139). To increase generalizability of the approach, different MRI sequences with varying view, matrix size, and field of view were used for SSL training (summarized in Table 1).

6As a practical application, we focused on segmenting joint effusion from the sagittal IW TSE (intermediate-weighted turbo spin echo) MRI which is becoming more important as osteoarthritis is increasingly recognized to have inflammatory components including joint effusion. Effusion regions were segmented by a trained MSK radiologist for a total of 31 scans using an interactive software developed in-house [10].

**Table 1.** Information of the deployed MRI sequences for SSL training, SSL evaluation and Segmentation (SAG = sagittal, COR= coronal, AX = Axial)

|  |  | 3D DESS | IW TSE | MPR |  | T1-3D FLASH | T1-THIGH | T2-MAP | MP Locator | STIR |
|---|---|---|---|---|---|---|---|---|---|---|
|  | **View** | SAG | SAG | COR | AX | COR | COR | AX | SAG | - | COR |
|  | **Body Part** | Knee | Knee | Knee | Knee | Knee | Knee | Knee | Knee | Knee | Hip |
|  | **Pixel size** $mm^2$ | 0.364 | 0.357 | 0.364 | 0.364 | 0.364 | 0.312 | 0.976 | 0.312 | 0.39 | 0.91 |
|  | **Matrix** | 384×384 | 444×448 | 384×384 | 384×384 | 384×384 | 512×512 | 512×256 | 384×384 | 512×512 | 512×512 |
| **SSL** | **Train** | Yes | Yes | Yes | Yes | No | Yes | Yes | Yes | No | No |
|  | **Eval.** | No | Yes | Yes | Yes | Yes | No | No | No | Yes | Yes |
|  | **Segmentation** | No | Yes | No | No | No | No | No | No | No | No |

### 4.3 Evaluation Criteria and Experiment Setup

Evaluation was performed for the underlying SSL and the target segmentation tasks. For the SSL task we computed similarity measures by comparing the original image ($I_{org}$), and recovered image ($I_{ssl}$) and the original image ($I_{org}$) and distorted image ($I_{dist}$), to determine the amount of lost information than SS-MRCNN can recover. The similarity measurements included: Structural Similarity Index (SSIM), Peak Signal to Noise Ratio (PSNR) and Cosine similarity (CS). The evaluation metrics used for the instance segmentation task included precision (= TP/(TP + FP)) and recall (= TP/(TP + FN)) for detection, as well as the Dice similarity score (= 2×TP/(2×TP +FP +FN)) for segmentation. For all metrics higher values indicate better performance.

To assess the effectiveness of SS-MRCNN in relation to the size of the pretraining input data, we performed pretraining in different setups in un-labeled dataset sizes: 1) Ultra large dataset (with 247K training data), and 2) large target dataset (28k dataset used for downstream task), 3) small target dataset (100 slices from target data).

To investigate effect of pretraining for segmentation task, we compared fine-tuning using the three SSL pretrained weights with MS-COCO weights and with no pretraining. Furthermore, we assessed performance of the network on training data size with the labeled dataset (sagittal IW TSE MRI scans) in three sizes, a) 700 slices from 23 scans, and b) 100 slices from only 3 scans c) 10 slices. For the validation and test purposes, one scan (23 slices) and 7 scans (200 slices) have been used, respectively.

### 4.4 Results and Discussion

**SSL Representation learning:** Results show higher similarity measures between the full-sized original image and the recovered image $(I_{org}, I_{ssl})$ in compared to similarity measures between the original image and distorted image $(I_{org}, I_{dist})$ ($I_{dist}$ with 6 distortions) pair. This improvement was more than 5 dB for PSNR, 3%-6% for SSIM, and 1%-4% for Cosine similarity (Table 2). This indicates that the SS-MRCNN framework recovered part of the lost information. Meaning the network learned dense representation and is generalizable since it recovered distorted images of unseen MRI sequences. Furthermore, as can be observed in Fig. 2, the network learned to localize and differentiate the distortions indicating that it learned semantic and geometric information.

**Table 2.** Similarity measures between the $(I_{org}, I_{dist})$, and the $(I_{org}, I_{ssl})$, and their difference $\Delta = SIMILARITY(I_{org}, I_{ssl}) - SIMILARITY(I_{org}, I_{dist})$. The * and ** sequences were not seen by SS-MRCNN training, and ** is from unseen MSK structure.

| Measures | SSIM | | | PSNR (dB) | | | CS | | |
|---|---|---|---|---|---|---|---|---|---|
| Sample Sequences | $(I_{org}, I_{dist})$ | $(I_{org}, I_{ssl})$ | $\Delta$ | $(I_{org}, I_{dist})$ | $(I_{org}, I_{ssl})$ | $\Delta$ | $(I_{org}, I_{dist})$ | $(I_{org}, I_{ssl})$ | $\Delta$ |
| SAG_IW_TSE | 0.79 | 0.84 | 0.05 | 19.50 | 27.13 | 7.63 | 0.79 | 0.81 | 0.02 |
| COR_IW_TSE | 0.76 | 0.83 | 0.07 | 14.34 | 20.99 | 6.65 | 0.87 | 0.89 | 0.02 |
| AX_MPR | 0.81 | 0.87 | 0.06 | 18.25 | 25.45 | 7.20 | 0.87 | 0.90 | 0.03 |
| COR_MPR* | 0.84 | 0.87 | 0.05 | 19.37 | 25.46 | 6.09 | 0.82 | 0.83 | 0.01 |
| MP LOCATOR* | 0.78 | 0.81 | 0.03 | 17.50 | 22.61 | 5.11 | 0.92 | 0.94 | 0.02 |
| COR T1 (Hip MRI)** | 0.79 | 0.83 | 0.04 | 18.58 | 24.13 | 5.55 | 0.76 | 0.80 | 0.04 |

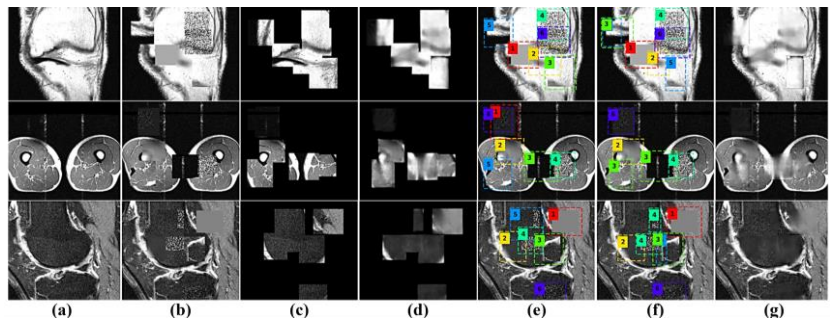

**Fig. 2.** Result of SSL task result, (a) original Image, (b) distorted Image, (c) ground truth mask, (d) predicted mask, (e) and (f) location and type of distortions in ground truth (e) and detected by SSL (f) [1- blank, 2- blurred, 3-mislocate, 4- salt & pepper noise, 5- rotate, 6- speckle noise], (g) recovered image.

**SSL Pretraining Effect:** Results of performing the segmentation and detection tasks in different settings (Table 3) shows, that pretraining with the SSL-pretraining as well as MS-COCO pretraining have a positive effect when comparatively small samples of labeled data are available for the downstream task and can boost the Dice score by 27%. However, comparing the two pre-training approaches shows that this positive effect is larger for SSL pre-training compared to MS-COCO pre-training as the size of the labeled dataset decreases. Table 3 clearly shows that as the labeled dataset size is reduced from 700 slices to 100 slices to 10 slices, the performance gap between the two pretraining methods increases. For the Dice similarity metric, this gap is 2%, 6%, and 21%, respectively. The Dice similarity metric for the MS-COCO pretraining is even lower when compared to the results when no pretraining is used for few-labeled data (10 slices). The visual analysis of the results (Fig. 3) supports the quantitative results. More precisely, it was found that the MS-COCO pretraining was not successful in defining correct edges around the effusion. In contrast, SSL-pretraining has a positive effect on recall as well, whereas high recall suggests that less effusions were missed by the network. However, the precision metric also shows that pretraining had no effect on reducing the number of false positives.

**Table 3.** Quantitative results of effusion segmentation.

| Pretraining Method | | No Pretrain | | | MS-COCO | | | SSL (Ultra large) | | | SSL (Large target) | | | SSL (small target) | |
|---|---|---|---|---|---|---|---|---|---|---|---|---|---|---|---|
| Labeled Data size | | 700 | 100 | 10 | 700 | 100 | 10 | 700 | 100 | 10 | 700 | 100 | 10 | 100 | 10 |
| Detection | Recall | 0.83 | 0.66 | 0.47 | 0.95 | 0.82 | 0.45 | **0.97** | 0.90 | 0.69 | 0.94 | 0.90 | 0.69 | 0.85 | 0.68 |
| | Precision | 0.51 | 0.44 | 0.44 | 0.36 | 0.53 | 0.53 | 0.47 | 0.45 | 0.41 | 0.51 | 0.44 | 0.42 | 0.27 | 0.41 |
| Segmentation | Dice | 0.54 | 0.51 | 0.46 | 0.73 | 0.65 | 0.40 | **0.78** | **0.71** | **0.61** | **0.74** | **0.71** | **0.69** | **0.61** | **0.60** |

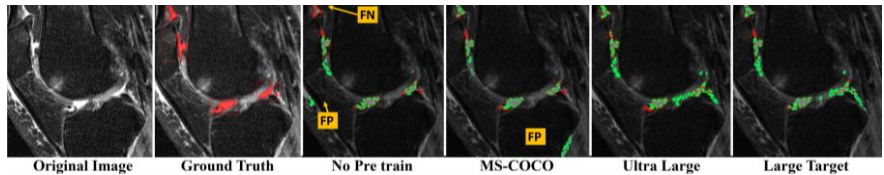

**Fig. 3.** Result of segmentation task for fine-tuning with 700 slices, green-dashes network segmentation, red dashes ground truth (FN=false negative, FP=false positive)

## 5    Conclusion

In this paper, we proposed SS-MRCNN which is a simple and effective SSL that learns both semantic and pixel level information for medical image segmentation. The SS-MRCNN uses a single network without requiring multi-view input, unlike CL methods that require multiple views of each training sample for positive pairs. The SS-MRCNN pre-training opens an avenue for extracting relevant information from large clinical MRI databases for which no annotations are available. The SS-MRCNN is a self-supervised pretraining method that is trained by adding distortion to images and learning to detect, classify, and restore these distortions. Effectiveness of SS-MRCNN pre-training was examined for effusion segmentation, which is challenging due to the class imbalance problem. The SS-MRCNN outperformed MS-COCO pretraining, and was shown to be effective for a medical image segmentation task with datasets of different sizes, even with small pretraining dataset, and few labeled data. By performing multiple pretext tasks, SS-MRCNN learns dense features in addition to semantic representations, which is crucial for pixel-wise prediction in segmentation and has not been explored in most of the previous studies. In compared to some other methods, the SS-MRCNN does not need to store extra information (e.g., negative pairs in CL), and does not need data augmentation, as it receives new samples in each iteration.  The SS-MRCNN is easily transferable to different domains making it suitable to use few target data for pretraining, and has the potential to increase performance of networks trained with few labels.

10In: Proceedings of the IEEE Computer Society Conference on Computer Vision and Pattern Recognition. pp. 2733–2742 (2018). https://doi.org/10.1109/CVPR.2018.00289.
16. Karimi, D. et al.: Transfer learning in medical image segmentation: New insights from analysis of the dynamics of model parameters and learned representations. Artif. Intell. Med. 116, 102078 (2021). https://doi.org/10.1016/J.ARTMED.2021.102078.
17. Khakzar, A. et al.: Towards Semantic Interpretation of Thoracic Disease and COVID-19 Diagnosis Models. Lect. Notes Comput. Sci. (including Subser. Lect. Notes Artif. Intell. Lect. Notes Bioinformatics). 12903 LNCS, 499–508 (2021). https://doi.org/10.1007/978-3-030-87199-4_47/FIGURES/5.
18. Lin, T.Y. et al.: Microsoft COCO: Common objects in context. In: Lecture Notes in Computer Science (including subseries Lecture Notes in Artificial Intelligence and Lecture Notes in Bioinformatics). pp. 740–755 Springer Verlag (2014). https://doi.org/10.1007/978-3-319-10602-1_48.
19. Liu, J. et al.: Prototypical Interaction Graph for Unsupervised Domain Adaptation in Surgical Instrument Segmentation. Lect. Notes Comput. Sci. (including Subser. Lect. Notes Artif. Intell. Lect. Notes Bioinformatics). 12903 LNCS, 272–281 (2021). https://doi.org/10.1007/978-3-030-87199-4_26/FIGURES/2.
20. Liu, Q. et al.: Federated Semi-supervised Medical Image Classification via Inter-client Relation Matching. Lect. Notes Comput. Sci. (including Subser. Lect. Notes Artif. Intell. Lect. Notes Bioinformatics). 12903 LNCS, 325–335 (2021). https://doi.org/10.1007/978-3-030-87199-4_31/FIGURES/2.
21. Marrakchi, Y. et al.: Fighting Class Imbalance with Contrastive Learning. Lect. Notes Comput. Sci. (including Subser. Lect. Notes Artif. Intell. Lect. Notes Bioinformatics). 12903 LNCS, 466–476 (2021). https://doi.org/10.1007/978-3-030-87199-4_44.
22. Ouyang, C. et al.: Self-supervision with Superpixels: Training Few-Shot Medical Image Segmentation Without Annotation. In: Lecture Notes in Computer Science (including subseries Lecture Notes in Artificial Intelligence and Lecture Notes in Bioinformatics). pp. 762–780 Springer Science and Business Media Deutschland GmbH (2020). https://doi.org/10.1007/978-3-030-58526-6_45.
23. Pathak, D. et al.: Context Encoders: Feature Learning by Inpainting. In: Proceedings of the IEEE Computer Society Conference on Computer Vision and Pattern Recognition. pp. 2536–2544 (2016). https://doi.org/10.1109/CVPR.2016.278.
24. Raghu, M. et al.: Transfusion: Understanding transfer learning for medical imaging. In: Advances in Neural Information Processing Systems. (2019).
25. Wang, T., Isola, P.: Understanding Contrastive Representation Learning through Alignment and Uniformity on the Hypersphere. 37th Int. Conf. Mach. Learn. ICML 2020. PartF16814, 9871–9881 (2020).
26. Wei, C. et al.: Masked Feature Prediction for Self-Supervised Visual Pre-Training.
27. Wu, Y. et al.: Federated Contrastive Learning for Volumetric Medical Image Segmentation. Lect. Notes Comput. Sci. (including Subser. Lect. Notes Artif. Intell. Lect. Notes Bioinformatics). 12903 LNCS, 367–377 (2021). https://doi.org/10.1007/978-3-030-87199-4_35/TABLES/2.
28. You, C. et al.: SimCVD: Simple Contrastive Voxel-Wise Representation Distillation for Semi-Supervised Medical Image Segmentation. IEEE Trans. Med. Imaging. (2022).